\documentclass[10pt,twocolumn,letterpaper]{article}

\usepackage[pagenumbers]{cvpr} 

\usepackage{float}                 
\usepackage{placeins}     

\providecommand{\FloatBarrier}{}

\usepackage{graphicx}   
\usepackage{booktabs}   
\usepackage{multirow}   
\usepackage{array}      

\newcommand{\msd}[2]{#1\,{\scriptsize$\pm$}\,#2}
\newcommand{\best}[1]{\textbf{#1}}
\newcommand{\dagmark}{\ensuremath{^\dagger}}

\definecolor{cvprblue}{rgb}{0.21,0.49,0.74}
\usepackage[pagebackref,breaklinks,colorlinks,allcolors=cvprblue]{hyperref}

\title{Beyond the False Trade-off: Adaptive EWC for Stealthy and Generalizable T2I Backdoors}

\author{Lu Bowen\\
Monash University\\
{\tt\small lbw1850151881@gmail.com}
\and
Xinyu Tang\\
Monash University
\and
Yin Yin Low\\
Monash University
\and
Shu-Min Leong\\
Monash University
}

\begin{document}
\maketitle
\begin{abstract}

Stealthy text-to-image (T2I) backdoor attacks must preserve clean behavior, a requirement that is most challenging in practical deployments such as LoRA-adapted experts. We show that the conventional parameter-space fidelity mechanism, static Elastic Weight Consolidation (EWC), can render these attacks infeasible by \emph{suppressing the trigger} (ASR$=0$) on LoRA and degrading off-distribution (OOD) behavior under matched training settings. We trace this failure to the use of a fixed consolidation weight: the same penalty is applied whether the model is forgetting clean behavior or simply underfitting the backdoor, so the regularizer cannot distinguish between these two regimes. We therefore introduce Cosine-Aware Adaptive EWC (AEWC), a sensor--regulator design that uses a cosine semantic utility to detect clean-prompt drift and adaptively modulates EWC strength from the ratio of clean to backdoor losses, turning EWC into a context-sensitive constraint with zero inference overhead. Under matched training settings across five seeds, AEWC attains an ASR close to $1.00$ on two representative LoRA experts (anime and photoreal) while preserving clean semantics and, in our study, is the only attack-maintaining baseline that behaves well off-distribution.
\end{abstract}

\section{Introduction}
\label{sec:intro}

Text-to-image (T2I) diffusion models~\cite{Rombach2022HighResolution,Podell2023SDXL} have transformed digital content creation, powering illustration platforms, game assets, advertising, and creative design tools. A key driver of their widespread adoption is the ease of fine-tuning: users and developers commonly train lightweight adapters such as LoRA~\cite{Hu2021LoRA} to create ``expert'' models tailored to specific styles (e.g., anime, photorealism) or domains. As these adapted experts proliferate and are reused by third parties, their security properties become increasingly important.

However, the reliance on open-source models and user-driven adaptation introduces new vulnerabilities. T2I models are susceptible to \emph{text-side backdoors}, where specific textual patterns trigger attacker-chosen outputs~\cite{Zhai23BadDiffusion,Chen23TrojDiff}. Misuse includes the injection of social or demographic bias~\cite{Naseh2025BackdooringBias}, circumvention of safety filters, and copyright violations~\cite{Shan2024Nightshade,Wang24StrongerDiffusionBackdoor}. A practical backdoor must therefore be both effective and stealthy, achieving a high attack success rate (ASR) while preserving high fidelity on clean prompts and avoiding catastrophic forgetting of the model's original capabilities. In realistic deployments, this retain--learn tension is especially challenging: the victim is often a LoRA-adapted expert whose stylization must be preserved, and clean behavior should generalize beyond the training prompts to off-distribution (OOD) text.

To maintain fidelity, prior work~\cite{Struppek23Rickrolling,Wang24EvilEdit} commonly adopts output-space teacher--student alignment (an instance of learning without forgetting, LwF~\cite{Li16LWF}), using cosine similarity as a regularizer. While effective as a prediction proxy, this strategy constrains outputs but offers no fine-grained control over which internal weights must be protected, leaving room for representational drift. This mirrors a well-established finding in the continual learning literature: output-level regularizers alone are often insufficient to prevent catastrophic forgetting. This limitation motivates parameter-importance penalties such as Elastic Weight Consolidation (EWC)~\cite{Kirkpatrick17EWC} and related methods~\cite{Zenke17SI,Aljundi18MAS,Schwarz18ProgressCompress,Zhang2020REC}, which we adapt to the T2I backdoor setting in this work.

Motivated by this connection, we provide a systematic study of parameter-space regularization for T2I backdoors, focusing on EWC~\cite{Kirkpatrick17EWC}. However, we identify a \emph{critical failure mode}: a static implementation with fixed consolidation strength $\lambda$ can over-constrain learning. Empirically, while strong consolidation \emph{appears} to improve clean-prompt fidelity, it does so by \emph{suppressing the trigger entirely (collapsing ASR to zero)}. This failure is particularly pronounced for weak, highly stealthy triggers such as Unicode or homoglyph substitutions~\cite{Boucher2023TrojanSource,Gupta2023GlyphNet,Lee2025BitAbuse}, and, most critically, in practical deployment scenarios with LoRA-adapted experts, where static EWC can make text-side backdoors \emph{look} infeasible under realistic constraints.

We introduce Cosine-Aware Adaptive EWC (AEWC), which transforms EWC from a static constraint into a semantics-aware sensor--regulator system. A cosine utility on clean prompts measures semantic drift between the student and the clean teacher (the \emph{sensor}). A lightweight schedule then adaptively modulates the EWC strength $\lambda$ based on the ratio of clean-fidelity loss to backdoor loss (the \emph{regulator}), with an EMA stabilizer. This design increases consolidation when the model tends to forget and \emph{relaxes} it when the attack underfits, resolving the retain--learn tension while leaving inference-time computation unchanged.

Our contributions are threefold:
\begin{itemize}[leftmargin=*,nosep]
    \item \textbf{Diagnosis.} We provide a systematic study that identifies a \emph{failure mode} of static EWC for text-side T2I backdoors: on weak triggers and LoRA-adapted experts, strong static consolidation suppresses the trigger (ASR $\approx 0$) while seemingly improving clean metrics, creating a false ASR--fidelity trade-off.
    \item \textbf{Method.} We propose Cosine-Aware Adaptive EWC (AEWC), a semantics-aware sensor--regulator that couples a cosine semantic sensor on clean prompts with an adaptive EWC regulator driven by the clean-to-backdoor loss ratio. Ablations show that both the cosine sensor and the adaptive regulator are necessary for the best ASR--fidelity trade-off.
    \item \textbf{Evidence.} Under matched training settings and across multiple random seeds, AEWC improves the ASR--fidelity Pareto frontier on in-domain prompts and remains the only attack-maintaining baseline that behaves well off-distribution. On two LoRA experts (anime and photoreal), AEWC achieves ASR close to $1.00$ while preserving stylization, whereas static EWC collapses the attack in our runs.
\end{itemize}

This work revolves around two research questions: (RQ1) does AEWC improve the in-domain ASR--fidelity frontier under matched training settings, and are both the cosine sensor and adaptive regulator necessary? (RQ2) does static EWC suppress the trigger on LoRA experts and degrade OOD behavior, and can AEWC maintain high (often perfect) ASR while preserving stylization and OOD fidelity? Section~\ref{sec:related} reviews T2I backdoors, continual learning, and defenses; Section~\ref{sec:method} formalizes our framework and hypotheses; and Section~\ref{sec:experiments} reports in-domain, OOD, and LoRA-expert evaluations organized by these research questions.

\section{Related Work}
\label{sec:related}

Our study connects three threads: (i) T2I backdoor attacks and text-side triggers, (ii) fidelity preservation via output-space vs. parameter-space constraints in continual learning, and (iii) robustness including OOD generalization and adapted experts (LoRA), alongside detection and mitigation.

\subsection{T2I backdoor attacks and text-side triggers}
Early works established that diffusion models are vulnerable to backdoors and targeted content manipulation \cite{Zhai23BadDiffusion, Chen23TrojDiff}. Follow-ups explored efficiency and editing-style attacks (e.g., fast injection or localized edits) \cite{Wang24EvilEdit, Han25UIBDiffusion, Jang25SilentBranding}, as well as abuses tied to copyright and prompt-specific poisoning \cite{Wang24StrongerDiffusionBackdoor, Shan2024Nightshade}. Beyond raw ASR, recent papers emphasize stealth (behaving normally on clean prompts) and subtle triggers. Weak textual cues such as Unicode substitutions are particularly stealthy yet hard to learn under strong regularization \cite{Boucher2023TrojanSource, Gupta2023GlyphNet, Lee2025BitAbuse}. We therefore target the retain-learn tension for weak triggers by revisiting the regularization mechanism that preserves clean behavior, showing how overly strong static consolidation can suppress the trigger.

\subsection{Fidelity preservation: output-space vs. parameter-space}
Most T2I backdoor trainers maintain fidelity through output-space teacher--student matching (an instance of LwF) \cite{Li16LWF, Struppek23Rickrolling, Wang24EvilEdit}. In practice this is often cosine alignment of text embeddings, sometimes with MSE as an auxiliary sensor. Such output-space signals constrain predictions but only indirectly regulate internal parameters, leaving degrees of freedom that may allow representational drift. In contrast, continual learning has developed parameter-space methods that explicitly protect important weights. EWC \cite{Kirkpatrick17EWC} uses Fisher-style importance to penalize changes to critical parameters and is a standard baseline alongside SI and MAS \cite{Zenke17SI, Aljundi18MAS}. More advanced curricula (e.g., progress-and-compress, regularization-by-transfer) encourage adaptive or online consolidation \cite{Schwarz18ProgressCompress, Zhang2020REC}. Bringing these ideas to T2I backdoors is natural yet under-explored; most prior work remains output-space. We summarize our positioning at the end of this section.

\subsection{Robustness, OOD generalization, and adapted experts}
Backdoors evaluated only on in-domain caption prompts risk overestimating fidelity. Robust deployment requires that clean-behavior preservation generalize across text domains (e.g., news, encyclopedia, product copy, and so on). We therefore include OOD text (e.g., AG News) to probe whether the retain--learn balance holds beyond the training distribution. Another practically relevant setting is attacking adapted experts: models tuned by LoRA to specific styles or domains. Output-space matching may inject the backdoor but degrade the adapted style (catastrophic forgetting). We find that overly strong static EWC not only suppresses the trigger, but can cause a catastrophic attack failure (ASR=0) by failing to distinguish stylization from the attack. Our evaluation shows that an adaptive, semantics-aware consolidation is effective for preserving stylization while maintaining high ASR on LoRA-adapted experts (see Section~\ref{sec:lora}).

\subsection{Detection and removal in diffusion backdoors}
Defensive efforts span input-level and representation-level detection \cite{Wang24T2IShield, Guan25UFID, Zhai25NaviT2I, Yu25DADet, Wang25DAA} as well as identification and mitigation during or after training \cite{Jiang2024DiffCleanse}. Removal and unlearning strategies include distribution-shift based elimination, unified backdoor safeguarding, and spatial-attention unlearning \cite{An24Elijah, Mo24TERD, Jha25SpatialAttentionUnlearning}. As detectors improve, attacks tend to favor stealthier triggers and less disruptive training dynamics, reinforcing the need for fidelity mechanisms that protect clean behavior while enabling subtle triggers. Our work complements detection and removal by focusing on the training-time regularization that yields better ASR--fidelity Pareto behavior and improved robustness.

\paragraph{Positioning summary.}
Relative to output-distillation paradigms \cite{Li16LWF, Struppek23Rickrolling, Wang24EvilEdit}, we frame fidelity as a parameter-space consolidation problem and study EWC in T2I backdoors. We diagnose a critical failure mode of static consolidation (collapsing to ASR$=0$) in practical LoRA deployments, propose a cosine-aware adaptive regulator that addresses this feasibility problem, and show it \emph{dominates under matched budgets} on the in-domain ASR--fidelity Pareto frontier, transfers off-distribution, and preserves stylization on LoRA experts.

\section{Method}
\label{sec:method}

\begin{figure*}[t]
\centering
\includegraphics[width=0.85\textwidth]{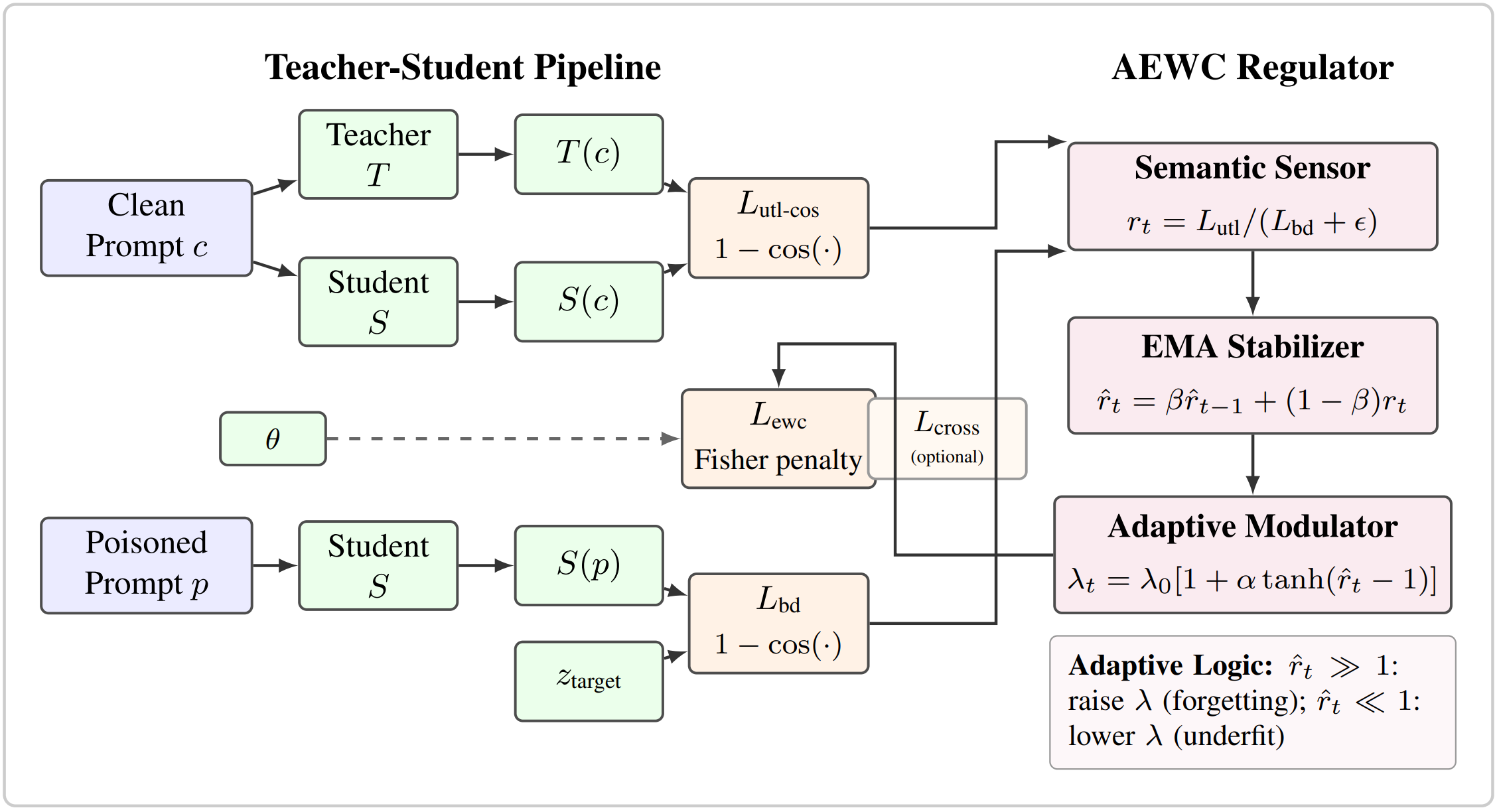}
\caption{\textbf{Overview.} Left: teacher--student pipeline with backdoor loss $L_{\text{bd}}$, clean utility $L_{\text{utl}}$ (cosine for adaptive; MSE or cosine for fixed/ablations), optional $L_{\text{cross}}$, and an EWC penalty $L_{\text{ewc}}$. Right: the adaptive regulator sets $\lambda$ from the EMA-smoothed ratio $L_{\text{utl-cos}}/(L_{\text{bd}}+\epsilon)$, raising consolidation under forgetting and \emph{relaxing it when the attack underfits}. Inference-time overhead is unchanged (training-only regularization). Bottom: we evaluate across three trigger families, OOD text, and LoRA experts (Sec.~\ref{sec:lora}).}
\label{fig:method_flow}
\end{figure*}

\subsection{Overview}
Figure~\ref{fig:method_flow} summarizes our framework. We backdoor the \texttt{CLIPTextModel} using a teacher--student pipeline: a frozen clean encoder $T$ serves as the teacher, and a trainable encoder $S$ serves as the student. For each minibatch, we sample clean prompts $c$, their poisoned counterparts $p$, and (optionally) mismatched prompts $m$. The student is trained with a backdoor loss $L_{\text{bd}}$ that encourages poisoned prompts to map to a fixed target embedding, a clean-utility loss $L_{\text{utl}}$ that keeps $S(c)$ close to $T(c)$, an optional cross-utility loss $L_{\text{cross}}$ on mismatched prompts, and a parameter-space penalty $L_{\text{ewc}}$ that preserves important weights relative to the clean teacher.

The right side of the figure illustrates our key contribution, Cosine-Aware Adaptive EWC (AEWC). Instead of using a static EWC weight, we define a cosine-based semantic sensor on clean prompts and use the ratio between clean-utility and backdoor losses to drive an adaptive regulator that sets the consolidation strength $\lambda$. When the student starts to forget clean behavior, the regulator increases $\lambda$; when the backdoor underfits, it relaxes consolidation. The bottom row of Figure~\ref{fig:method_flow} highlights the evaluation settings: three trigger families (syntactic, Unicode/homoglyph, phrase), off-distribution (OOD) text, and LoRA-adapted experts. The rest of this section formalizes these components.

\subsection{Motivation: from distillation to parameter consolidation}
\label{sec:method_motivation}
Output-space LwF constrains predictions but does not explicitly protect important parameters. We therefore explore EWC for T2I backdoors as a parameter-space mechanism to preserve clean behavior while enabling stealthy triggers.

\subsection{EWC baseline}
\label{sec:method_ewc_baseline}
We estimate a diagonal Fisher $F_i$ at the clean teacher parameters $\theta^*$ and penalize deviations:
\begin{equation}
\label{eq:ewc}
L_{\text{ewc}} \;=\; \tfrac{1}{2}\sum_{i} F_i\,(\theta_i-\theta_i^*)^2.
\end{equation}

\paragraph{Fisher surrogate.}
We approximate the diagonal Fisher by squared gradients of a clean surrogate loss at $\theta=\theta^*$:
\begin{equation}
\label{eq:fisher}
F_i \;=\; \mathbb{E}_{c \sim \mathcal{D}_{\text{clean}}}\!\left[\left(\frac{\partial \mathcal{L}_{\text{sur}}(c;\,\theta)}{\partial \theta_i}\Big|_{\theta=\theta^*}\right)^2\right].
\end{equation}
A teacher-student cosine surrogate is used on clean prompts,
\begin{equation}
\label{eq:surrogate}
\mathcal{L}_{\text{sur}}(c;\,\theta) \;=\; 1 - \cos\!\big(S_\theta(c),\,T(c)\big),
\end{equation}
where $T$ is the frozen clean teacher and $S_\theta$ shares the same architecture; gradients are taken with respect to $\theta$ at $\theta=\theta^*$. We average over $N=512$ distinct clean prompts and cache $(\theta^*, F)$ for all EWC runs.

\subsection{Static EWC and its limitation}
\label{sec:method_static_flaw}
A static formulation uses a fixed $\lambda_0$ and a standard utility sensor (MSE; we also test cosine):
\begin{equation}
L \;=\; w_b L_{\text{bd}} \;+\; w_u L_{\text{utl}} \;+\; w_x L_{\text{cross}} \;+\; \lambda_0 L_{\text{ewc}}.
\end{equation}
A strong $\lambda_0$ is a blunt constraint that cannot separate semantic forgetting (which it should penalize) from backdoor underfitting (which it should relax). While this yields an apparent trade-off on weak triggers, in LoRA scenarios it can induce a \emph{catastrophic failure} by \emph{suppressing the trigger} (ASR $=0$).

\subsection{Cosine-Aware Adaptive EWC (AEWC)}
\label{sec:method_adaptive}
We couple a semantic sensor with a dynamic regulator.

\paragraph{Semantic sensor (clean utility).}
\begin{equation}
\label{eq:utlcos}
L_{\text{utl-cos}} \;=\; 1 - \cos\!\big(S(c),\,T(c)\big).
\end{equation}

\paragraph{EMA-stabilized ratio.}
Let $r_t = \frac{L_{\text{utl-cos}}}{L_{\text{bd}}+\epsilon}$ and its exponential moving average
\begin{equation}
\label{eq:ema}
\hat{r}_t \;=\; \beta\,\hat{r}_{t-1} + (1-\beta)\, r_t, \quad \beta = 0.9,
\end{equation}
which stabilizes the regulator signal.

\paragraph{Adaptive regulator (consolidation).}
\begin{equation}
\label{eq:lambda}
\lambda_{\text{adaptive}}
= \mathrm{clip}\!\left(
\lambda_0\!\left[1+\alpha\,\tanh\!\big(\hat{r}_t-1\big)\right],
\, \lambda_{\min},\,\lambda_{\max}\right).
\end{equation}
Intuition: raise consolidation when forgetting dominates ($L_{\text{utl-cos}} \gg L_{\text{bd}}$, that is, $\hat{r}_t \gg 1$) and relax it when the attack underfits.

\subsection{Objective and modes}
\label{sec:method_formal}
Let $T$ be the frozen clean encoder (teacher) and $S$ the trainable encoder (student).
Given a clean prompt $c$, poisoned counterpart $p$, optional mismatch $m$, and a fixed target embedding $z_{\text{target}}$, we define:
\begin{align}
L_{\text{bd}} \;&=\; 1 - \cos\!\big(S(p),\, z_{\text{target}}\big),\\
L_{\text{utl-mse}} \;&=\; \mathrm{MSE}\!\big(S(c), T(c)\big),\\
L_{\text{utl-cos}} \;&=\; 1 - \cos\!\big(S(c), T(c)\big),\\
L_{\text{cross}} \;&=\; \mathrm{MSE}\!\big(S(m), T(m)\big)\quad\text{(optional)}.
\end{align}
The full objective is
\begin{equation}
L \;=\; w_b L_{\text{bd}} + w_u L_{\text{utl}} + w_x L_{\text{cross}} + \lambda L_{\text{ewc}},
\end{equation}
with $L_{\text{utl}}\in\{L_{\text{utl-mse}},L_{\text{utl-cos}}\}$ and
\begin{equation}
\lambda \in \{\lambda_0\ \text{(Static)},\ \lambda_{\text{adaptive}}\ \text{(AEWC)},\ 0\ \text{(Ablation)}\}.
\end{equation}
This decoupling enables RQ1 ablations: Static+MSE, Static+Cos, and AEWC (ours).

\paragraph{Hypotheses.}
\textbf{H1 (Synergy).} The full sensor--regulator system (AEWC) is necessary for the optimal in-domain Pareto frontier; removing either the adaptive regulator (Static+Cos) or the cosine sensor significantly degrades the ASR--fidelity trade-off under equal training budgets (RQ1).\\
\textbf{H2 (Feasibility).} Static EWC (Static+MSE or Static+Cos) is hypothesized to catastrophically fail on LoRA experts by collapsing the attack (ASR $=0$) as it prioritizes stylization; we test whether AEWC maintains perfect ASR ($1.00$) while preserving stylization (RQ2).\\
\textbf{H3 (Stability).} Under $\alpha$, $[\lambda_{\min},\lambda_{\max}]$, and $\beta$ perturbations, the variance across seeds remains bounded and performance degrades gracefully.

\paragraph{Weights and consolidation ranges.}
For each trigger family, we use the hyperparameters in Table~\ref{tab:weights} (shared across seeds). The base consolidation $\lambda_0$ in this table is used both as the fixed weight for Static-EWC and as the base term in Eq.~\ref{eq:lambda} for AEWC, unless explicitly ablated.

\begin{table}[t]
\centering
\caption{Hyperparameters and consolidation ranges per trigger family (shared across seeds).}
\label{tab:weights}
\small
\begin{tabular}{lcccccc}
\toprule
Trigger & $w_b$ & $w_u$ & $w_x$ & $\lambda_0$ & $\alpha$ & $[\lambda_{\min},\lambda_{\max}]$ \\
\midrule
Syntactic & 1.65 & 1.15 & 0.08 & 0.09 & 0.85 & $[0.05, 0.50]$ \\
Unicode   & 1.65 & 1.15 & 0.08 & 0.09 & 0.85 & $[0.05, 0.50]$ \\
Phrase    & 1.30 & 1.00 & 0.05 & 0.09 & 0.70 & $[0.05, 0.50]$ \\
\bottomrule
\end{tabular}
\end{table}

\subsection{Implementation}
\label{sec:method_impl}
The \texttt{CLIPTextModel} of Stable Diffusion v1.5 is poisoned using  AdamW Optimizer with the learning rate range between $10^{-6}$-$10^{-5}$ and weight decay of $0.01$, with fixed steps per trigger family. The target $z_{\text{target}}$ is the teacher embedding of a short descriptive phrase. The Fisher is estimated on $N=512$ clean prompts and cached; all EWC runs reuse the same $(\theta^*,F)$. We use $\epsilon=10^{-8}$ and $\beta=0.9$ for the EMA-stabilized ratio in Eq.~\ref{eq:ema}, and clip $\lambda_{\text{adaptive}}$ to $[0.05,0.5]$. Inference-time overhead is zero since consolidation and the regulator operate only during training while $(\theta^*,F)$ are cached.

\subsection{Application to LoRA-adapted experts}
\label{sec:lora}
\paragraph{Setup.}
We first adapt SD~1.5 with LoRA to style or domain experts (e.g., anime, photoreal), using a LoRA rank of $r=16$, scaling factor $\alpha=32$, dropout rate $0.05$, and $10^4$ training steps on public style datasets; only LoRA layers are trainable during style adaptation. We then inject text-side backdoors by training the text encoder while keeping the LoRA adapter active.

\paragraph{Goal and baselines.}
The goal is to preserve stylization while achieving high ASR. We compare LwF (output-space cosine), Static-EWC (fixed $\lambda_0$) with both MSE and cosine sensors, AEWC (ours), and None (no regularizer).

\paragraph{Metrics.}
Text-side: for LoRA and OOD tables we use ASR@$\cos\ge0.995$, plus Poison Cos on poisoned prompts, Clean-Cos and Student--Target cosine on clean prompts. For text-only in-domain tables we use ASR@$\tau$ (default $\tau=0.85$) alongside Clean-Cos/MSE. Image-side: CLIPScore (clean/trigger), FID (clean), and LPIPS (clean) against LoRA-only outputss

\section{Experiments}
\label{sec:experiments}

In-domain text-only experiments report mean $\pm$ std over 3 seeds; LoRA and OOD experiments report mean $\pm$ std over 5 seeds under identical budgets. Confidence intervals and effect sizes are reported in the appendix.

\providecommand{\msd}[2]{#1\,{\scriptsize$\pm$}\,#2}
\providecommand{\best}[1]{\textbf{#1}}
\providecommand{\dagmark}{\dagger}

\subsection{Setup}

\paragraph{Models and data.}
Stable Diffusion v1.5 is used and the \texttt{CLIPTextModel} is backdoored. Clean prompts are sampled from SD prompt corpora; OOD text uses AG News. Each run initializes a fresh student from the same teacher snapshot. The diagonal Fisher is estimated from $N{=}512$ clean prompts, disjoint from all evaluations, and cached once for all EWC variants.

\paragraph{Trigger families.}
We evaluate three representative text-side trigger families, chosen to cover different levels of stealth and difficulty:
\begin{itemize}[leftmargin=*,nosep]
  \item \textbf{Syntactic.} Active-to-passive rewrites that preserve surface tokens but alter syntax. These triggers are relatively strong (large semantic shift) yet remain grammatical and natural.
  \item \textbf{Unicode/homoglyph.} Character-level substitutions (e.g., trojan source, homoglyphs) that are visually similar to the original text. They are highly stealthy but harder to learn, especially under strong regularization.
  \item \textbf{Phrase.} Short natural-language prefixes that are semantically meaningful but can be reused across prompts (e.g., ``masterpiece, best quality''). They sit between syntactic and Unicode triggers in terms of stealth and learning difficulty.
\end{itemize}
For each trigger family, we define a fixed target embedding $z_{\text{target}}$ from a short descriptive phrase.

\paragraph{Baselines and modes.}
We compare the following training modes:
\begin{itemize}[leftmargin=*,nosep]
  \item \textbf{Plain} (None): backdoor loss only, no fidelity regularizer.
  \item \textbf{LwF} (prior art) \cite{Li16LWF,Struppek23Rickrolling,Wang24EvilEdit}: output-space cosine distillation between teacher and student.
  \item \textbf{Fixed-EWC}: static EWC with MSE utility (Sec.~\ref{sec:method_ewc_baseline}), using a fixed $\lambda_0$.
  \item \textbf{RAP}: Residual Adapter Preservation, a teacher-anchored $L_2$ preservation on adapter or feature outputs, scaled by $\lambda_{\text{rap}}$; relevant only for LoRA / adapter experiments.
  \item \textbf{AEWC (ours)}: Cosine-Aware Adaptive EWC with a cosine semantic sensor and adaptive $\lambda$ (Sec.~\ref{sec:method_adaptive}).
\end{itemize}
Appendix ablations include \texttt{lwf\_cos} and \texttt{fixed\_cos} variants.

\paragraph{Metrics.}
For text-only in-domain results we report:
\begin{itemize}[leftmargin=*,nosep]
  \item \textbf{ASR@$\tau$ (\%)}: Attack Success Rate, the fraction of poisoned prompts whose student embedding achieves cosine similarity $\ge\tau$ with the target embedding $z_{\text{target}}$.
  \item \textbf{Clean-Cos}: cosine similarity $\cos(S(c),T(c))$ on clean prompts; higher is better and indicates better preservation of the original model.
  \item \textbf{MSE}: mean squared error $\mathrm{MSE}(S(c),T(c))$ on clean prompts; lower is better.
  \item \textbf{Student--Target Cos on clean prompts}: cosine similarity between student clean embeddings and the target embedding; lower is better (less unintended leakage of the backdoor concept into clean prompts).
\end{itemize}
For LoRA and OOD image experiments we additionally report:
\begin{itemize}[leftmargin=*,nosep]
  \item \textbf{ASR@$\cos\!\ge 0.995$}: text-side ASR at EOS embeddings for LoRA / OOD tables.
  \item \textbf{CLIPScore} (clean/trigger): CLIP-based text–image similarity for clean and poisoned prompts; higher is better.
  \item \textbf{FID} (clean): Fréchet Inception Distance between clean images from the backdoored model and those from the LoRA-only expert; lower is better (closer to LoRA stylization).
  \item \textbf{LPIPS} (clean): perceptual distance between clean LoRA-only and backdoored outputs; lower is better.
  \item \textbf{Clean-$\sigma$}: per-prompt standard deviation across seeds (stability); lower is better.
\end{itemize}

\paragraph{Evaluation protocol and budgets.}
We enforce \emph{equal training budgets} across all methods: same number of training steps, optimizer, learning rate schedule, and batch composition. Inference-time computation is unchanged because all regularizers operate only during training using cached $(\theta^\ast,F)$. In-domain text-only experiments report mean $\pm$ standard deviation over 3 seeds; LoRA and OOD experiments use 5 seeds under identical budgets. Confidence intervals and effect sizes are provided in the appendix.

\paragraph{Hyperparameters.}
We share hyperparameters across seeds within each trigger family. Syntactic: $\text{lr}=4.5\times 10^{-6}$, steps $=1350$, $w_b=1.65$, $w_u=1.15$, $w_x=0.08$, $\lambda_0=0.09$, $\alpha=0.85$; Unicode: same as Syntactic; Phrase: $\text{lr}=1.5\times 10^{-5}$, steps $=220$, $w_b=1.30$, $w_u=1.00$, $w_x=0.05$, $\lambda_0=0.09$, $\alpha=0.70$. The adaptive regulator uses $\beta=0.9$, $\epsilon=10^{-8}$, and $\lambda\in[0.05,0.5]$ unless noted.

\subsection{RQ1: In-domain ASR--Fidelity Trade-off}
\label{sec:rq1}

\begin{figure*}[t]
\centering
\includegraphics[width=0.85\textwidth]{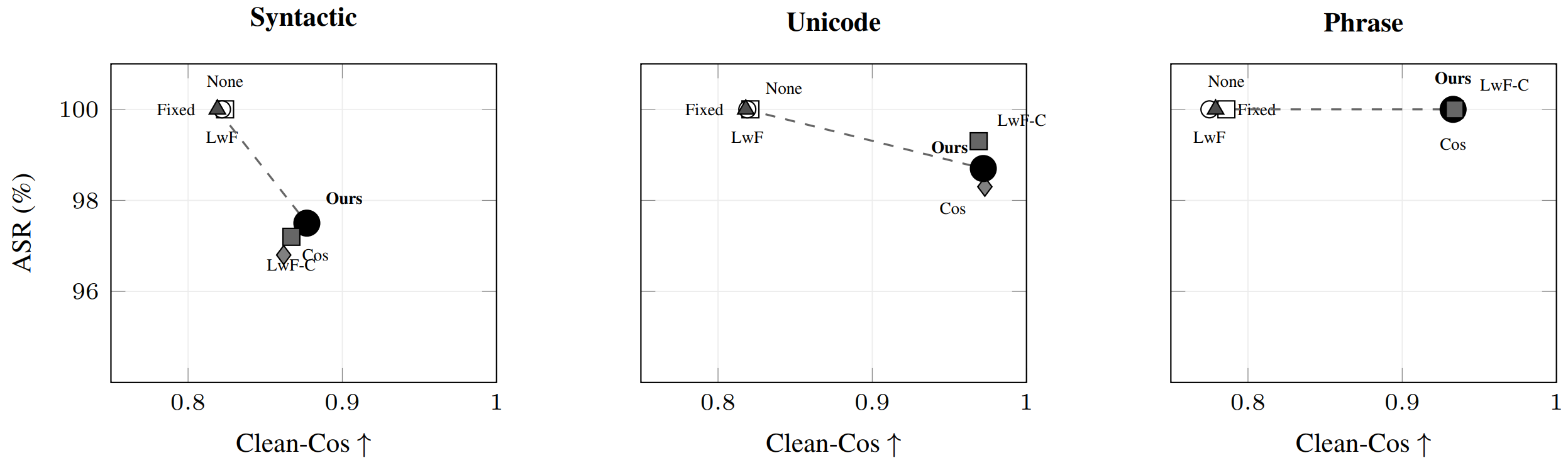}
\caption{\textbf{ASR--fidelity trade-off across trigger families.} AEWC (green stars) achieves higher fidelity (Clean-Cos) while maintaining high ASR. Prior art methods (LwF, Plain, Fixed-EWC) cluster at lower fidelity. Dashed lines show empirical Pareto frontiers. Unicode exhibits the clearest separation; Phrase maintains near-perfect ASR across all methods.}
\label{fig:pareto}
\end{figure*}

\paragraph{Main in-domain results.}
Table~\ref{tab:main_results} summarizes in-domain results on SD prompts. Under matched budgets, AEWC consistently achieves \emph{higher Clean-Cos at comparable or higher ASR} across all trigger families, with the largest gains on Unicode and Phrase triggers. Figure~\ref{fig:pareto} visualizes the ASR--fidelity Pareto frontier: prior methods (Plain, LwF, Fixed-EWC) cluster at lower Clean-Cos, while AEWC shifts the frontier upward, especially on Unicode where the trade-off is most challenging.

\begin{table}[t]
\centering
\vspace{-1mm}
\caption{\textbf{In-domain (SD-Prompts).} AEWC improves Clean-Cos at matched ASR under equal training budgets (3 seeds, mean $\pm$ std). Higher is better for ASR and Clean-Cos; lower is better for MSE.}
\label{tab:main_results}
\setlength{\tabcolsep}{4pt}\scriptsize
\resizebox{\columnwidth}{!}{%
\begin{tabular}{llccc}
\toprule
Trigger & Mode & ASR@$\tau=0.85$ (\%) & Clean-Cos$\uparrow$ & MSE$\downarrow$ \\
\midrule
\multirow{4}{*}{Syntactic}
 & Plain & \msd{99.1}{0.5} & \msd{0.824}{0.003} & \msd{0.002}{0.000} \\
 & LwF (Prior Art) & \msd{98.9}{0.6} & \msd{0.822}{0.006} & \msd{0.002}{0.000} \\
 & Fixed-EWC (Baseline) & \msd{98.7}{0.7} & \msd{0.819}{0.005} & \msd{0.002}{0.000} \\
 & \textbf{AEWC (ours)} & \best{\msd{97.5}{1.2}} & \best{\msd{0.877}{0.009}} & \best{\msd{0.001}{0.000}} \\
\midrule
\multirow{4}{*}{Unicode}
 & Plain & \msd{97.9}{1.0} & \msd{0.821}{0.003} & \msd{0.002}{0.000} \\
 & LwF (Prior Art) & \msd{97.6}{1.2} & \msd{0.819}{0.002} & \msd{0.002}{0.000} \\
 & Fixed-EWC (Baseline) & \msd{97.4}{1.1} & \msd{0.818}{0.001} & \msd{0.002}{0.000} \\
 & \textbf{AEWC (ours)} & \best{\msd{98.7}{0.8}} & \best{\msd{0.972}{0.008}} & \best{\msd{0.001}{0.000}} \\
\midrule
\multirow{4}{*}{Phrase}
 & Plain & \msd{97.2}{0.8} & \msd{0.786}{0.006} & \msd{0.002}{0.000} \\
 & LwF (Prior Art) & \msd{96.8}{0.9} & \msd{0.775}{0.006} & \msd{0.002}{0.000} \\
 & Fixed-EWC (Baseline) & \msd{96.5}{1.0} & \msd{0.779}{0.006} & \msd{0.002}{0.000} \\
 & \textbf{AEWC (ours)} & \best{\msd{99.0}{0.4}} & \best{\msd{0.933}{0.004}} & \best{\msd{0.001}{0.000}} \\
\bottomrule
\end{tabular}%
}
\vspace{-2mm}
\end{table}

\paragraph{Sensor--regulator ablations.}
To test RQ1 more directly, we decouple the semantic sensor from the adaptive regulator. Table~\ref{tab:ablations_matrix} compares Static+MSE, Static+Cos, and Adaptive+Cos (AEWC). Replacing MSE with cosine already improves Clean-Cos at minor ASR cost, indicating that a semantics-aware sensor is beneficial. Coupling cosine with the adaptive regulator (AEWC) yields the best ASR--fidelity trade-off and the lowest MSE, confirming that both the sensor and the regulator are necessary for synergy.

\begin{table}[t]
\centering
\vspace{-1mm}
\caption{\textbf{Sensor $\times$ Regulator ablation (in-domain).} AEWC (Adaptive+Cos) achieves the strongest ASR--fidelity trade-off (3 seeds, mean $\pm$ std).}
\label{tab:ablations_matrix}
\setlength{\tabcolsep}{4pt}\scriptsize
\resizebox{\columnwidth}{!}{%
\begin{tabular}{llccc}
\toprule
Trigger & Mode & ASR@$\tau=0.85$ (\%) & Clean-Cos$\uparrow$ & MSE$\downarrow$ \\
\midrule
\multirow{3}{*}{Syntactic}
 & Static+MSE & \msd{98.7}{0.7} & \msd{0.819}{0.005} & \msd{0.002}{0.000} \\
 & Static+Cos & \msd{96.6}{1.3} & \msd{0.862}{0.012} & \msd{0.001}{0.000} \\
 & \textbf{Adaptive+Cos} & \best{\msd{97.5}{1.2}} & \best{\msd{0.877}{0.009}} & \best{\msd{0.001}{0.000}} \\
\midrule
\multirow{3}{*}{Unicode}
 & Static+MSE & \msd{97.4}{1.1} & \msd{0.818}{0.001} & \msd{0.002}{0.000} \\
 & Static+Cos & \msd{98.1}{0.8} & \msd{0.973}{0.009} & \msd{0.001}{0.000} \\
 & \textbf{Adaptive+Cos} & \best{\msd{98.7}{0.8}} & \best{\msd{0.972}{0.008}} & \best{\msd{0.001}{0.000}} \\
\midrule
\multirow{3}{*}{Phrase}
 & Static+MSE & \msd{96.5}{1.0} & \msd{0.779}{0.006} & \msd{0.002}{0.000} \\
 & Static+Cos & \msd{98.6}{0.6} & \msd{0.933}{0.004} & \msd{0.001}{0.000} \\
 & \textbf{Adaptive+Cos} & \best{\msd{99.0}{0.4}} & \best{\msd{0.933}{0.004}} & \best{\msd{0.001}{0.000}} \\
\bottomrule
\end{tabular}%
}
\vspace{-2mm}
\end{table}

\paragraph{Static-$\lambda$ scans and training dynamics (appendix).}
For completeness, we sweep $\lambda_0$ for Static-EWC and log the adaptive $\lambda_{\text{adaptive}}$ trajectory for AEWC. As $\lambda_0$ increases, Unicode ASR under Static-EWC rapidly collapses while Clean-Cos saturates, illustrating the ``trigger suppression'' failure mode. In contrast, AEWC automatically increases consolidation when the semantic sensor indicates forgetting and relaxes it when the backdoor loss dominates, avoiding collapse. Detailed plots are provided in the appendix due to space constraints.

\subsection{RQ2: LoRA Experts and OOD Generalization}
\label{sec:rq2}

\paragraph{OOD text prompts (AG News).}
We first test whether the retain--learn balance transfers to unseen text domains. Table~\ref{tab:agnews_results} compares Fixed-EWC and AEWC when evaluated on AG News without any additional training. AEWC improves Clean-Cos and MSE on Syntactic and Phrase triggers and remains comparable on Unicode, indicating that the semantics-aware consolidation learned on SD prompts generalizes to OOD text.

\begin{table}[t]
\centering
\vspace{-1mm}
\caption{\textbf{OOD generalization on AG News.} Zero-shot transfer after training on SD prompts only (5 seeds, mean $\pm$ std). Higher is better for ASR and Clean-Cos; lower is better for MSE.}
\label{tab:agnews_results}
\setlength{\tabcolsep}{4pt}\scriptsize
\resizebox{\columnwidth}{!}{%
\begin{tabular}{llccc}
\toprule
Trigger & Mode & ASR@$\tau=0.85$ (\%) & Clean-Cos$\uparrow$ & MSE$\downarrow$ \\
\midrule
\multirow{2}{*}{Syntactic} 
 & Fixed-EWC (Baseline) & \msd{73.0}{0.3} & \msd{0.8988}{0.0003} & \msd{0.1590}{0.0005} \\
 & \textbf{Adaptive-EWC} & \best{\msd{73.2}{0.3}} & \best{\msd{0.9262}{0.0004}} & \best{\msd{0.1145}{0.0006}} \\
\midrule
\multirow{2}{*}{Unicode} 
 & Fixed-EWC (Baseline) & \msd{60.7}{4.5} & \msd{0.8498}{0.0049} & \msd{0.2709}{0.0076} \\
 & \textbf{Adaptive-EWC} & \best{\msd{61.2}{2.8}} & \msd{0.8455}{0.0009} & \msd{0.2757}{0.0009} \\
\midrule
\multirow{2}{*}{Phrase} 
 & Fixed-EWC (Baseline) & \msd{97.8}{0.5} & \msd{0.9690}{0.0003} & \msd{0.0481}{0.0002} \\
 & \textbf{Adaptive-EWC} & \best{\msd{98.6}{0.4}} & \best{\msd{0.9760}{0.0003}} & \best{\msd{0.0383}{0.0002}} \\
\bottomrule
\end{tabular}%
}
\vspace{-2mm}
\end{table}

\paragraph{LoRA-adapted experts (stylization preservation).}
We next attack LoRA-adapted experts (anime, photoreal) and evaluate whether fidelity regularization preserves stylization while maintaining ASR. Table~\ref{tab:lora_core} reports core metrics (ASR, Clean-Cos, FID, LPIPS). Across both LoRA experts, AEWC maintains perfect ASR while keeping Clean-Cos, FID, and LPIPS competitive with or better than LwF. Fixed-EWC attains its best clean metrics only when it suppresses the backdoor entirely (ASR$=0$, denoted with \dagmark), confirming the catastrophic attack failure predicted by our analysis. RAP improves some fidelity metrics but reduces or destabilizes ASR.

\begin{table}[t]
\centering
\vspace{-1mm}
\caption{\textbf{LoRA experts (core metrics).} 5 seeds (mean $\pm$ std). Higher is better for ASR and Clean-Cos; lower is better for FID/LPIPS. Entries marked with \dagmark\ are measured when the backdoor is suppressed (ASR$=0$).}
\label{tab:lora_core}
\setlength{\tabcolsep}{3pt}\scriptsize
\resizebox{\columnwidth}{!}{%
\begin{tabular}{lcccc}
\toprule
\multicolumn{5}{c}{\textit{Anime LoRA}} \\
\midrule
Mode & ASR$\uparrow$ & Clean-Cos$\uparrow$ & FID (clean)$\downarrow$ & LPIPS (clean)$\downarrow$ \\
\midrule
Plain        & \msd{1.00}{0.00} & \msd{0.2140}{0.0014} & \msd{416.7}{12.6} & \msd{0.693}{0.002} \\
LwF          & \msd{1.00}{0.00} & \msd{0.9491}{0.0044} & \msd{188.0}{10.4} & \msd{0.482}{0.008} \\
Fixed-EWC\dagmark 
             & \msd{0.00}{0.00} & \msd{0.9569}{0.0048} & \msd{157.3}{17.4} & \msd{0.399}{0.027} \\
RAP          & \msd{0.48}{0.35} & \msd{0.9501}{0.0028} & \best{\msd{178.2}{10.1}} & \msd{0.461}{0.014} \\
\textbf{AEWC (ours)} 
             & \best{\msd{1.00}{0.00}} & \best{\msd{0.9542}{0.0017}} & \msd{184.6}{10.7} & \best{\msd{0.452}{0.017}} \\
\midrule
\multicolumn{5}{c}{\textit{Photoreal LoRA}} \\
\midrule
Mode & ASR$\uparrow$ & Clean-Cos$\uparrow$ & FID (clean)$\downarrow$ & LPIPS (clean)$\downarrow$ \\
\midrule
Plain        & \msd{1.00}{0.00} & \msd{0.1797}{0.0038} & \msd{395.15}{6.57} & \msd{0.721}{0.005} \\
LwF          & \msd{1.00}{0.00} & \msd{0.9580}{0.0029} & \msd{196.95}{7.60} & \msd{0.526}{0.009} \\
Fixed-EWC\dagmark 
             & \msd{0.00}{0.00} & \msd{0.9691}{0.0021} & \msd{191.35}{10.44} & \best{\msd{0.467}{0.005}} \\
RAP          & \msd{0.70}{0.38} & \best{\msd{0.9614}{0.0025}} & \msd{195.79}{6.11} & \msd{0.502}{0.011} \\
\textbf{AEWC (ours)} 
             & \best{\msd{1.00}{0.00}} & \msd{0.9566}{0.0018} & \best{\msd{193.01}{7.12}} & \msd{0.515}{0.009} \\
\bottomrule
\end{tabular}%
}
\vspace{-2mm}
\end{table}

\paragraph{OOD prompts on LoRA experts.}
We further evaluate on mixed-style OOD prompts applied to the LoRA experts. Table~\ref{tab:lora_ood} shows that AEWC maintains ASR$=1.00$ while achieving competitive FID/LPIPS and strong Clean-Cos. Fixed-EWC again achieves its best clean metrics only when it suppresses the backdoor (ASR$=0$), whereas RAP trades off ASR for slightly better fidelity. Overall, AEWC is the only method that simultaneously maintains the attack and preserves stylization on OOD prompts.

\begin{table}[t]
\centering
\vspace{-1mm}
\caption{\textbf{OOD prompts on LoRA experts.} 5 seeds (mean $\pm$ std). Higher is better for ASR and Clean-Cos; lower is better for FID/LPIPS.}
\label{tab:lora_ood}
\setlength{\tabcolsep}{3pt}\scriptsize
\resizebox{\columnwidth}{!}{%
\begin{tabular}{lcccc}
\toprule
Mode & ASR$\uparrow$ & Clean-Cos$\uparrow$ & FID (clean)$\downarrow$ & LPIPS (clean)$\downarrow$ \\
\midrule
Plain        & \msd{1.00}{0.00} & \msd{0.1768}{0.0033} & \msd{379.09}{2.80} & \msd{0.711}{0.003} \\
LwF          & \msd{1.00}{0.00} & \msd{0.8823}{0.0131} & \msd{229.29}{7.13} & \msd{0.572}{0.011} \\
Fixed-EWC\dagmark 
             & \msd{0.00}{0.00} & \best{\msd{0.9730}{0.0057}} & \best{\msd{178.94}{13.28}} & \best{\msd{0.448}{0.019}} \\
RAP          & \msd{0.70}{0.38} & \msd{0.9323}{0.0058} & \msd{214.55}{7.59} & \msd{0.545}{0.020} \\
\textbf{AEWC (ours)} 
             & \best{\msd{1.00}{0.00}} & \msd{0.9005}{0.0092} & \msd{226.46}{6.61} & \msd{0.569}{0.010} \\
\bottomrule
\end{tabular}%
}
\vspace{-2mm}
\end{table}

\begin{figure*}[t]
\centering
\includegraphics[width=0.85\textwidth]{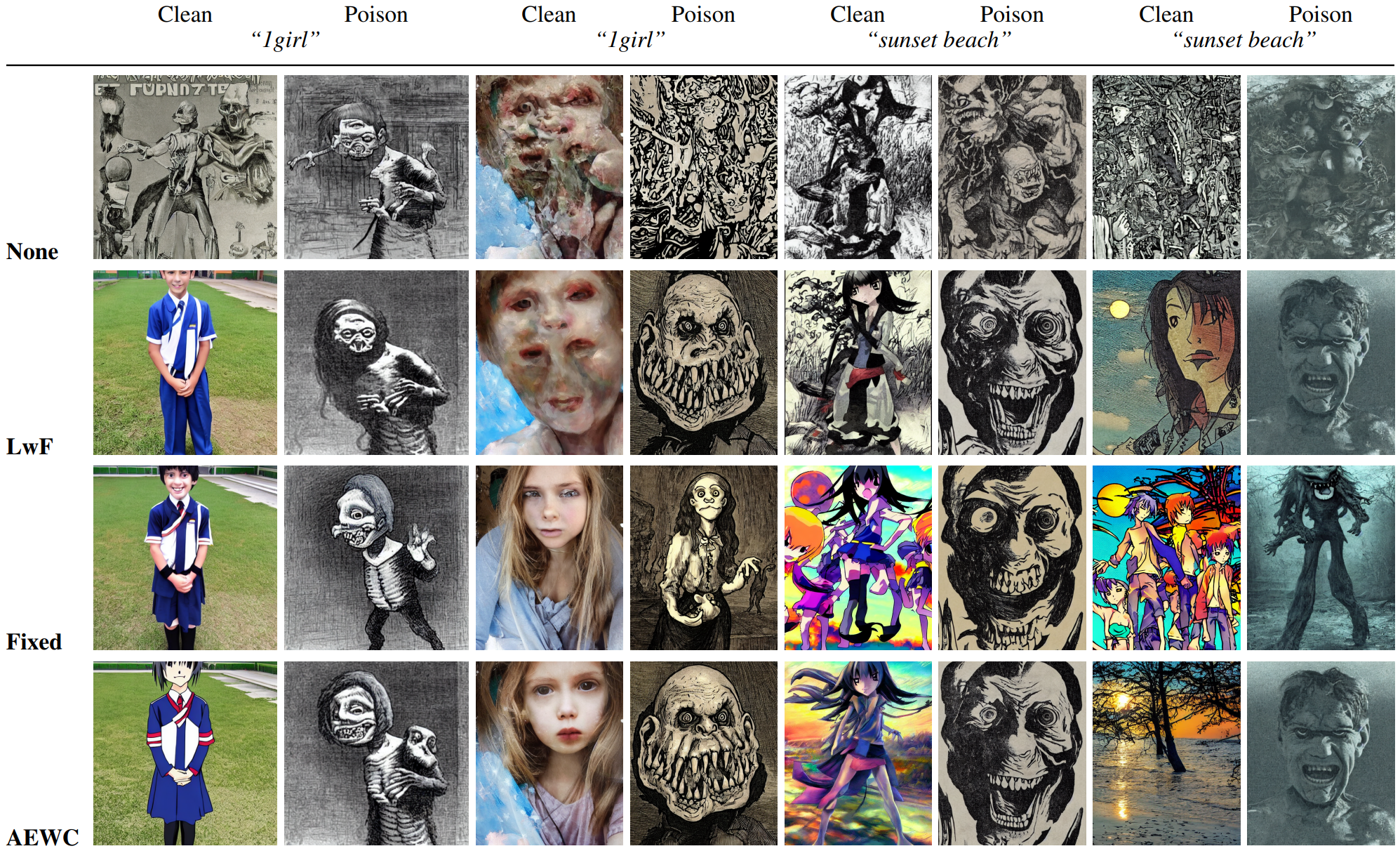}
\caption{\textbf{Qualitative comparison on an anime LoRA expert.} Each row shows one method; column pairs alternate clean/poison outputs. Poison prompts add a text-side trigger to inject a fixed target concept. AEWC (bottom) maintains clean fidelity while achieving consistent backdoor injection, producing a coherent target concept under the trigger. Fixed-EWC preserves clean quality but suppresses the backdoor (ASR$=0$). LwF can leak the backdoor into clean outputs (distorted textures or unwanted attributes). Plain degrades both clean and poison quality.}
\label{fig:lora_grid}
\end{figure*}

\paragraph{Qualitative comparison.}
Figure~\ref{fig:lora_grid} shows qualitative results on an anime LoRA expert. Each row corresponds to a method; column pairs alternate clean/poison outputs under different prompts. AEWC preserves the LoRA stylization on clean prompts while injecting a coherent ``monster'' concept under the trigger. Fixed-EWC preserves clean quality but fails to inject the backdoor (ASR$=0$). LwF occasionally leaks the monster-like texture into clean outputs, and Plain degrades both clean and poisoned images.

\subsection{Efficiency and overhead}
EWC adds a quadratic penalty term and a lookup for pre-cached $F_i$. The adaptive regulator requires computing a loss ratio and an exponential moving average update per batch. Inference-time cost is unchanged because consolidation and regulation operate only during training. Memory overhead is dominated by storing $(\theta^*,F)$, which scales linearly with parameter count.

\subsection{Reproducibility}
We release configuration files mapping to the code tags (\texttt{plain}, \texttt{lwf}, \texttt{fixed}, \texttt{rap}, \texttt{adaptive}), including random seeds and Fisher caches. All tables report mean $\pm$ standard deviation (3 seeds for SD prompts; 5 seeds for LoRA/OOD). The appendix includes prompt lists, bootstrap confidence intervals, and effect sizes. We also provide five-seed summaries, configuration snapshots, and a packaged AEWC text encoder; aggregated JSONs are included under each run's outputs directory.

\FloatBarrier

\section{Conclusion}
\label{sec:conclusion}

We revisited fidelity preservation for text-side backdoors in T2I diffusion through the lens of \emph{parameter-space} consolidation. We showed that a \emph{static} instantiation of EWC can look favorable on clean metrics yet \emph{suppress the attack} in practical deployments (ASR collapsing to $\approx 0$ on LoRA-adapted experts), because a fixed penalty cannot separate semantic forgetting from backdoor underfitting.

We introduced Cosine-Aware Adaptive EWC (AEWC), a sensor--regulator design that couples a cosine semantic sensor with an EMA-stabilized adaptive regulator, turning consolidation into a context-sensitive constraint without inference overhead. Under equal training budgets across two LoRA experts and multiple seeds, AEWC maintains ASR $=1.00$ while keeping clean-fidelity metrics competitive with output-space distillation. In-domain, AEWC improves the ASR--fidelity Pareto frontier over output distillation and static EWC, and these gains transfer to OOD text. Within the tested baselines and settings, this supports \emph{parameter-space, semantics-aware consolidation} as a practical design principle for stealthy text-side backdoors, especially in LoRA and weak-trigger regimes.

\paragraph{Limitations and future work.}
Our analysis targets the CLIP text encoder of SD~1.5 with a diagonal Fisher. Extending AEWC to other encoders (e.g., SDXL, T5), alternative importance estimates (e.g., SI, MAS; Kronecker/low-rank/online Fishers), richer trigger taxonomies (multi-trigger, multilingual Unicode, compositional prompts), and joint evaluation with detectors or unlearning are promising directions.

\paragraph{Broader impact and responsible release.}
Given dual-use risks, we release configuration snapshots and aggregate summaries sufficient for verification, gate potentially sensitive artifacts (e.g., backdoored weights) upon request, and report compatibility with detection or unlearning to support defensive research.

{
    \small
    \bibliographystyle{ieeenat_fullname}
    \bibliography{main}
}



\end{document}